\documentclass{article}

\usepackage[final]{nips_2016}

\usepackage[utf8]{inputenc} 
\usepackage[T1]{fontenc}    
\usepackage{hyperref}       
\usepackage{url}            
\usepackage{booktabs}       
\usepackage{amsfonts}       
\usepackage{nicefrac}       
\usepackage{microtype}      

\title{Minimally Naturalistic Artificial Intelligence}

\author{
  Steven Hansen \\
  Department of Psychology\\
  Stanfored University\\
  Stanford, CA 94303 \\
  \texttt{sshansen@stanford.edu} \\
}

\begin{document}

\maketitle

\section{Introduction}

The rapid advancement of machine learning techniques has re-energized research into general artificial intelligence. While the idea of domain-agnostic meta-learning is appealing, this emerging field must come to terms with its relationship to human cognition and the statistics and structure of the tasks humans perform. The position of this article is that only by aligning our agents' abilities and environments with those of humans do we stand a chance at developing general artificial intelligence (GAI).

A broad reading of the famous “No Free Lunch” theorem[1] is that there is no universally optimal inductive bias or, equivalently, bias-free learning is impossible. This follows from the fact that there are an infinite number of ways to extrapolate data, any of which might be the one used by the data generating environment; an inductive bias prefers some of these extrapolations to others, which lowers performance in environments using these adversarial extrapolations. We may posit that the optimal GAI is the one that maximally exploits the statistics of its environment to create its inductive bias; accepting the fact that this agent is guaranteed to be extremely sub-optimal for some alternative environments. This trade-off appears benign when thinking about the environment as being the physical universe, as performance on any fictive universe is obviously irrelevant. But, we should expect a sharper inductive bias if we further constrain our environment.  Indeed, we implicitly do so by defining GAI in terms of accomplishing that humans consider useful. One common version of this is need the for “common-sense reasoning”, which implicitly appeals to the statistics of physical universe as perceived by humans [2].

\section{The CommAI Environment}
The CommAI environment[3] is both too broad and too narrow. It is too broad in that the tasks need not bear in relation to those encountered by humans, and are untethered to the statistics that lie therein. For example, imagine that a CommAI agent is trying to follow written driving direction to reach allocation in a simulated car. Most of the time the teacher gives standard instructions akin to those of a human. But sometimes, directions are given under the assumption that the world is governed by non-Euclidean geometry (e.g. "drive until these parallel streets intersect"). An optimal agent should spend time trying to infer the geometry governing the teacher’s directions. But such an agent is woefully sub-optimal in the real world, as directions given under these beliefs are far too rare to warrant such exploratory behavior. \emph{The variation in the CommAI environment doesn’t match the variation in the real environment.} 

This criticism appears to be addressed by appealing to meta-learning or learning-to-learn. For example, by first exploring the space of possible teachers, and then forming an exploration strategy based on the empirical variation. But this merely pushes the problem to a higher level of abstraction. We can imagine an environment whereby the \emph{type} of variation in the set of teachers does not align with reality, such as teachers’ whose geometric believes change by the day of the week. An optimal agent would thus have to track the statistics of the teachers’ beliefs conditioned by the day of the week in order to infer the appropriate exploration strategy, which would clearly lead to spurious correlations in a naturalistic setting.

The CommAI environment also too narrow, in that it throws out naturalistic structure in the input and output representations in the pursuit of simplicity. While the bit stream passed between the the teacher and the agent can send image data, its one dimensional nature means stripping it of its two-dimensional structure. While this could be recovered by the agent, such a problem is non-trivial and would likely slow progress on its ultimate task, which is problematic since one of the environment’s primary goals is to strip away extraneous preprocessing in order to focus efforts on general learning principles. The authors attempt to address this criticism by the suggested use of object detection systems to convert image data into a format with a similarity structure akin to linguistic data, but this ignores the fact that much of the recent progress in machine learning has been due to systems that adapt their representations to the ultimate objective function, foregoing the modularization necessitated by an object detection system [4].

None of this is to say that the CommaAI environment won't be a useful tool in the pursuit of general artificial intelligence. Indeed, the high barrier to entry imposed by vision-based environments is probably the reason why more work isn’t done on higher level tasks like instruction taking. Separating these projects may accelerate progress in both areas, as the reasons behind unsuccessful modeling attempts can be more readily diagnosed. However, we must be cognizant that we’ve made this split for pragmatic reasons rather than philosophical ones.

\section{The Case of Reward Inference}

A concrete case where generic and minimally naturalistic approaches to artificial intelligence diverge is the problem of goal inference. While reward is in some sense the simplest possible feedback, there are many cases where even a reward signal is hard to come by[5]. Indeed, social learning appears to be one of these cases[6]. Instead, we must frame our agent's motivations in terms of a latent intention that the teacher has in mind, but can only communicate indirectly (e.g. sending reward, verbal instructions). The most generic way to handle such a situation is by taking the reward signal at face value and treat other indications of latent intention as additional state information. This was the approach taken in the work of Branavan et al[7] (though they did not have to handle the issue of imprecise rewards). In trying to solve a video-game (Civilization 2) via model-based reinforcement learning, they utilized the written instructions in the game manual to augment the state-space and focus the search. 

A more psychologically informed approach can be seen in the work of Ullman et al, where the task is to explain the behavior of other agents in the environment. One could take the generic approach and try and find a mapping between observations of the other agents to the correct explanation the behavior of other agents was understood by inverse planning on a model of each agent's behavior. Not only did this approach predict human performance on this task, related work has shown that this approach is also a computationally viable way to handle imprecise reward signals[6].

Having an agent model its teacher is very difficult, as the agent is likely to only observe the teacher is circumscribed, unrepresentative situations (e.g. when the teacher is about to teach something). However, if the teacher is sufficiently similar to the agent, the agent can use a model of its own behavior as a stand-in for a teacher model. One of the overarching assumptions of GAI is that we want to be the ones giving instructions; we want to be the teacher. The growing body of work on human learning suggests that in order to do so tractably, we should make a deliberate effort to make the inductive biases of our agent as close to ours as possible by training in naturalistic situations. The extent to which this constraint is weighed against others, such as the computational resources required for simulating an environment, is an open question, but some minimal commitment to naturalism is necessary.

\section*{References}
\small

[1] Wolpert, D. H., \& Macready, W. G. (1997). No free lunch theorems for optimization. IEEE transactions on evolutionary computation, 1(1), 67-82.

[2] Minsky, M. L., Singh, P., \& Sloman, A. (2004). The St. Thomas common sense symposium: designing architectures for human-level intelligence. AI Magazine, 25(2), 113.

[3] Mikolov, T., Joulin, A., \& Baroni, M. (2015). A roadmap towards machine intelligence. arXiv preprint arXiv:1511.08130.

[4] Mnih, V., Kavukcuoglu, K., Silver, D., Rusu, A. A., Veness, J., Bellemare, M. G., ... \& Petersen, S. (2015). Human-level control through deep reinforcement learning. Nature, 518(7540), 529-533.

[5] Loftin, R., Peng, B., MacGlashan, J., Littman, M. L., Taylor, M. E., Huang, J., \& Roberts, D. L. (2016). Learning behaviors via human-delivered discrete feedback: modeling implicit feedback strategies to speed up learning. Autonomous Agents and Multi-Agent Systems, 30(1), 30-59.

[6] Ullman, T., Baker, C., Macindoe, O., Evans, O., Goodman, N., \& Tenenbaum, J. B. (2009). Help or hinder: Bayesian models of social goal inference. In Advances in neural information processing systems (pp. 1874-1882).

[7] Branavan, S. R. K., Silver, D., \& Barzilay, R. (2011, June). Learning to win by reading manuals in a Monte-Carlo framework. In Proceedings of the 49th Annual Meeting of the Association for Computational Linguistics: Human Language Technologies-Volume 1 (pp. 268-277). Association for Computational Linguistics.

\end{document}